\PassOptionsToPackage{hyphens}{url}

\documentclass[journal]{IEEEtran} 
\IEEEoverridecommandlockouts
\usepackage{cite}
\usepackage{amsmath,amssymb,amsfonts}
\usepackage{algorithmic}
\usepackage{graphicx}
\usepackage{textcomp}
\usepackage{xcolor}
\usepackage{hyperref}
\usepackage{makecell}

\def\BibTeX{{\rm B\kern-.05em{\sc i\kern-.025em b}\kern-.08em
    T\kern-.1667em\lower.7ex\hbox{E}\kern-.125emX}}
\begin{document}

\title{Agentic AI for Mobile Network RAN Management and Optimization\\
}

\author{
    \IEEEauthorblockN{Jorge Pellejero}
    , \IEEEauthorblockN{Luis A. Hernández Gómez}
    , \IEEEauthorblockN{Luis Mendo Tomás}, \IEEEauthorblockN{Zoraida Frias Barroso}
    \\
    \IEEEauthorblockA{\textit{IPTC, ETSI de Telecomunicación, Universidad Politécnica de Madrid, Spain}}\\
    \IEEEauthorblockA{Corresponding author: jorge.pellejero@alumnos.upm.es }
}

\maketitle

\begin{abstract}
    Agentic AI represents a new paradigm for automating complex systems by using Large AI Models (LAMs) to provide human-level cognitive abilities with multimodal perception, planning, memory, and reasoning capabilities. This will lead to a new generation of AI systems that autonomously decompose goals, retain context over time, learn continuously, operate across tools and environments, and adapt dynamically. The complexity of 5G and upcoming 6G networks renders manual optimization ineffective, pointing to Agentic AI as a method for automating decisions in dynamic RAN environments. However, despite its rapid advances, there is no established framework outlining the foundational components and operational principles of Agentic AI systems nor a universally accepted definition.

    This paper contributes to ongoing research on Agentic AI in 5G and 6G networks by outlining its core concepts and then proposing a practical use case that applies Agentic principles to RAN optimization. We first introduce Agentic AI, tracing its evolution from classical agents and discussing the progress from workflows and simple AI agents to Agentic AI. Core design patterns—reflection, planning, tool use, and multi-agent collaboration—are then described to illustrate how intelligent behaviors are orchestrated. These theorical concepts are grounded in the context of mobile networks, with a focus on RAN management and optimization. A practical 5G RAN case study shows how time-series analytics and LAM–driven agents collaborate for KPI-based autonomous decision-making.
\end{abstract}



\section{Introduction}
With the proliferation of 5G, communication networks are becoming increasingly complex due to dense deployments, diverse service-level requirements, and high heterogeneity in Radio Access Network (RAN) configurations. The transition towards 6G is expected to further amplify these dynamics, placing even greater demands on network management, operational control, and optimization. In this context, traditional network management approaches—heavily dependent on rule-based scripts, static thresholds, and manual interventions—are very inefficient. To meet the challenges of next-generation networks, there is a pressing need for intelligent, adaptive, and scalable frameworks capable of managing the lifecycle of network services dynamically.

In order to augment the autonomy and efficiency of 5G advanced networks, intent-driven management emerges as a key enabler, allowing operators to define desired outcomes without specifying how they should be achieved. Implementing intent-based systems requires intelligent agents capable of semantic interpretation, contextual awareness, and decision-making under uncertainty—capabilities that are not fully realized by traditional automation techniques.

The research in \cite{habib2025} explores the use of Generative AI (GenAI) where Large Language Models (LLMs) and Retrieval-Augmented Generation (RAG) techniques are adapted for intent validation and decision-making in RAN management. 

Beyond LLMs, the evolution to Large AI Models (LAMs), which integrate LLMs, Large Vision Models (LVMs), Large Multimodal Models (LMMs), and Large Reasoning Models (LRM), represent an advancement for future intelligent communications. Nevertheless, GenAI and LAMs alone present important limitations to fulfill the high AI demands of current complexities of communication systems \cite{jiang2025survey}. To address these limitations, the emerging Agentic AI paradigm has recently arisen as a potential solution for future intelligent communications \cite{jiang2025tutorial}.

AI agents, which utilize LAM as central controller, can provide advanced capabilities towards future Artificial General Intelligence (AGI). Compared to AI agents, Agentic AI represents a more advanced approach that integrates large-scale pretrained models with planning, memory, and reasoning capabilities, allowing systems to autonomously decompose goals, execute actions across tools and contexts, and adapt dynamically \cite{jiang2025tutorial}.

While recent literature provides comprehensive overviews of Agentic AI \cite{jiang2025tutorial}, real-world deployments remain scarce and underexplored. This paper aims to bridge that gap by translating theoretical advancements into a concrete application: the use of Agentic AI for RAN optimization in mobile networks. Our contribution is twofold: first, we clarify the conceptual progression from classical agents to modern Agentic AI, and we examine the modular design patterns that underpin intelligent behaviors in current systems; second, we operationalize these ideas in a real-world RAN scenario. Specifically, we demonstrate how Agentic AI, when integrated with time-series analytics and language model–driven agents, enables autonomous KPI-based monitoring and decision-making in RAN management. 

The remainder of the paper is organized as follows. Section~\ref{sec:prin} introduces the foundations of AI workflows and agents, leading to a detailed exploration of Agentic AI principles. This section establishes the conceptual evolution from classical agents to modern autonomous systems, covering features such as memory integration, planning, goal decomposition, and reflective reasoning. Section~\ref{sec:patterns} reviews core design patterns in Agentic AI, including reflection, planning, tool use, and multi-agent collaboration, offering a modular view of how intelligent behaviors are orchestrated in contemporary frameworks. Section~\ref{sec:agenticRAN} discusses the main capabilities of Agentic AI for RAN Management and Optimization. References to recent industrial initiatives from leading telecom operators and vendors aiming at demonstrating the applicability across scenarios are given, and main technical and operational challenges for deployment are presented. Section~\ref{sec:casestudy} presents a practical case study on the use of Agentic AI for KPI-based monitoring in 5G RAN environments. This study showcases how time-series analytics and language model-driven agents can collaborate to support autonomous decision-making. Finally, Section~\ref{sec:conclu} summarizes the main contributions of this work.


\section{Agentic AI principles}\label{sec:prin}

Classical symbolic and rule-based AI was typically characterized by reactive or deliberative architectures operating under predefined rules and logic-based reasoning frameworks. These agents functioned in narrowly defined environments, relying on explicit representations of world states and deterministic planning to perform tasks such as pathfinding, decision-tree evaluation, and expert system inference. With advances in computational capacity and learning algorithms, agent architectures evolved to include probabilistic reasoning, reinforcement learning, and multi-agent coordination, enabling more adaptive behaviors in dynamic settings. A major shift followed with the integration of LLMs and GenAI techniques, producing agents capable of understanding natural language, generating context-aware actions, and interfacing with external tools. Modern agents now employ foundation models for perception, planning, and decision-making, greatly expanding their scope and autonomy.

The latest stage in this evolution is marked by the emergence of Agentic AI, which unifies generative capabilities with agent-based planning, memory, and goal management. These systems extend beyond conventional AI agents, and as discussed in \cite{sapkota2025}, Agentic AI represents a substantial step toward general-purpose autonomy, facilitating multi-step decision-making, adaptive tool use, and lifelong learning in open-ended environments.

Although the term Agentic AI was initially used informally—``there is no widely accepted definition of ‘AI agent’” \cite{casper2025}—recent research has begun to define it more precisely. Zhang et al. \cite{zhang2025networking} describe Agentic AI as involving autonomous agents that ``can perceive, reason, act, and continuously learn from their environments,” while Casper et al. \cite{casper2025} refer to foundation-model-based systems enhanced with planning, memory, and tool-use scaffolds, capable of executing complex tasks with limited human oversight. In this context, Agentic AI can be understood as an autonomous-agent paradigm in which advanced pre-trained models, equipped with reasoning and long-term planning, self-direct multi-step actions to pursue specified goals \cite{zhang2025networking, casper2025}. Sapkota et al. \cite{sapkota2025} add that such systems incorporate persistent memory, multi-agent collaboration, and autonomous decision-making, supporting scalable and context-aware behavior. Acharya et al. \cite{acharya2025} further note their applicability across domains such as healthcare, robotics, and network management.

\begin{table*}[t]
\caption{Comparison of Workflows, AI Agents, and Agentic AI}
\label{tab:agentic_comparison}
\centering
\small 
\renewcommand{\arraystretch}{1.5}
\begin{tabular}{|>{\raggedright\arraybackslash}p{3cm}|
                >{\raggedright\arraybackslash}p{4.2cm}|
                >{\raggedright\arraybackslash}p{4.2cm}|
                >{\raggedright\arraybackslash}p{4.2cm}|}
\hline
\textbf{Aspect} & \textbf{Workflows} & \textbf{AI Agents} & \textbf{Agentic AI} \\
\hline
Definition & Structured sequences of operations or reasoning steps & Systems that perceive, decide, and act based on goals and workflows & Goal-directed systems using foundation models for autonomous reasoning and adaptation \\
\hline
Autonomy & Not autonomous; driven by predefined rules and algorithms & Limited:  act within constrained task scopes bounded by predefined logic & High; can create, adapt, and coordinate workflows dynamically \\
\hline
Architecture & Composed of modules like prompting, retrieval, memory, tool use & Integrate workflows, sensors, planners, and memory (if any) & Unified architecture with planning, memory, reflection, and tool interfaces \\
\hline
Memory and\newline Adaptation & No persistent memory; reused patterns & Typically lack long-term memory & Include persistent memory and reflective planning loops; support long-horizon adaptation \\
\hline
Use of Foundation Models & Not required; may involve LLMs as tools & May use LLMs for perception or response generation & Core component; LAMs used for perception, reasoning, and planning \\
\hline
Goal Handling & Encoded implicitly in the sequence & Execute fixed goals or policies & Interpret abstract goals and plan actions to achieve them \\
\hline
Workflow Management & Defined manually or learned; static structure & Instantiates workflows at runtime from a predefined set. & Constructs and adapts workflows based on context \\
\hline
Scalability and\newline Flexibility & Limited to context of definition & Narrowly scoped; domain-specific & Designed for general-purpose use across domains with minimal retraining \\

\hline
\end{tabular}
\end{table*}


This paper explores how Agentic AI supports the development of autonomous, goal-driven agents capable of managing complex RAN configurations under dynamic and uncertain conditions with minimal human input. While GenAI systems like LLMs excel at generating context-aware outputs, they remain reactive and lack long-term goal orientation. Agentic AI features overcome these limitations providing a suitable foundation for intent-driven network management and automated optimization.

However, besides the important benefits of Agentic AI, current industrial experiences show that translating high-level human intelligence into orchestrated multi-agent ecosystems is not an easy task. These experiences demonstrate that many successful AI systems can be engineered through low-level workflows that model well-defined tasks. When extra adaptability is required, they can be extended with narrowly scoped agents, and fully Agentic architectures are only advisable when flexibility and model-driven decision-making are needed at scale \cite{anthropic2024}. Consequently, an incremental approach starting with low-level workflows, introducing specialized agents as necessary, and progressing toward multi-agents is an excellent strategy to progress towards the Agentic AI vision.

To clarify the conceptual progression from workflows to AI agents and Agentic AI, Table~\ref{tab:agentic_comparison} summarizes their distinguishing characteristics across dimensions such as autonomy, architecture, memory, goal handling, and scalability. Workflows represent modular, reusable patterns for orchestrating reasoning and action, often manually designed or learned through interaction \cite{anthropic2024}. These are instantiated by AI agents—goal-driven systems that dynamically select and execute workflows while maintaining internal state and responding to context. Agentic AI extends this model by integrating foundation models, persistent memory, and reflective planning into a unified architecture capable of decomposing complex goals and adapting behavior over time with minimal human input. 


\section{Agentic AI design patterns}\label{sec:patterns}

An Agentic AI design pattern refers to a reusable architectural or behavioral structure that defines how components of an intelligent agent interact to support autonomy, reasoning, planning, and goal execution. These patterns help formalize best practices in constructing complex AI systems by promoting scalability, modularity, and predictability—important properties for developing reliable autonomous systems. They also serve as conceptual tools for system designers integrating the capabilities of LAMs, APIs, planning modules, and memory components. The taxonomy of these patterns continues to evolve, shaped by both empirical deployments and theoretical analysis. Main core design patterns are briefly described in the following subsections (more details and practical considerations can be found in \cite{webweb}).

\subsection{Reflection pattern}\label{sec:reflec}
The reflection pattern enables an agent to evaluate and revise its outputs through self-assessment and iterative refinement. Applied to LLMs, it introduces a dynamic loop in which the model reviews its own outputs, identifies weaknesses, and adjusts accordingly. This process enhances accuracy, efficiency, and contextual relevance. Despite its simplicity, the Reflection pattern improves reliability by combining self-evaluation with selective tool use, equipping LLMs with greater adaptability to evolving or complex tasks. When incorporated into agentic systems, it strengthens autonomy and decision quality with minimal human intervention. For example, an agent monitoring network KPIs may generate an initial hypothesis about performance degradation, then revise its conclusion after re-evaluating with updated metrics or incorporating peer comparisons.

\subsection{Tool use pattern}
Tool Use refers to the ability of LLMs to interface with external tools
to perform complex tasks beyond text generation. This extends the model’s capabilities by allowing it to retrieve real-time information, execute calculations, or interact with databases, thus overcoming the limitations of static pre-trained knowledge. Tool Use supports function calling where LLMs generate structured commands to invoke tools and process results within the same workflow. One application of this pattern is the use of a statistical analysis tool to detect potential changes in KPI time-series data.

\subsection{Planning pattern}\label{sec:planning}

Planning involves the autonomous breakdown of complex tasks into structured, manageable steps. It is closely tied to reasoning, as the model must assess objectives, evaluate possible actions, and determine an optimal sequence to achieve a goal. This dynamic strategy formulation allows the AI to adapt its plan based on available tools, task constraints, and evolving conditions. Through Planning, the agent exhibits flexibility and creative problem-solving, especially in uncertain or open-ended scenarios. Rather than executing static instructions, the AI reasons through the task, selecting and coordinating actions to fulfill the objective. For example, in RAN optimization, an agent might detect a coverage issue, then devise a plan that includes identifying affected cells, analyzing traffic and interference patterns, retrieving historical KPIs, and suggesting parameter adjustments or beam reconfiguration to restore performance. While this autonomy enhances capability, it also introduces some unpredictability in behavior and outcomes.

\subsection{Multi-agent collaboration}\label{sec:collab}
In scenarios requiring distributed expertise or scalability, this pattern coordinates multiple specialized agents to perform subtasks. Multi-agent collaboration is a foundational design pattern in agentic AI that divides complex tasks into subtasks handled by specialized agents, each assigned a distinct role such as coder, planner, or reviewer. These agents can be instantiated either by prompting a single LLM with tailored instructions or by using multiple LLMs, each optimized for its function. This setup mirrors human organizational teams and parallels multi-threaded computing, offering efficiency and precision through concurrent and specialized execution. Beyond task division, agents collaborate via message passing and coordinated reasoning, adapting dynamically to complex workflows. Frameworks like AutoGen, CrewAI, and LangGraph operationalize this pattern, enabling developers to build scalable, role-based AI teams. While the approach boosts output quality and system modularity, it also introduces coordination overhead and requires careful management of inter-agent interactions. One application of this pattern in automated RAN management involves a coordination agent delegating performance analysis to one agent, historical data retrieval to another, and policy validation to a third, with all agents contributing to a unified optimization decision.

\subsection{Use of design patterns}

These core patterns—reflection, planning, tool-use and multi-agent collaboration—are not isolated but interdependent design principles. They are increasingly integrated in modern LAM-based systems, where modular frameworks allow flexible orchestration of perception, interaction, and execution. Their integration improves adaptability, transparency, and reproducibility by encapsulating reasoning strategies that can be reused and evaluated independently. As the field matures, these design patterns are likely to form the basis for building the next generation Agentic AI and to be codified into open standards, enabling more interpretable and robust AI across domains—from software development to human-machine collaboration in embodied settings.


\section{Agentic AI for RAN management and optimization}\label{sec:agenticRAN}


\subsection{Agentic AI Capabilities for RAN}

For intent-based RAN management and automated network optimization, Agentic AI provides collaborative intelligent agents equipped with the capabilities presented in Section~\ref{sec:patterns}. These agents can interpret high-level intents (e.g., minimize latency, balance load, reduce energy consumption), monitor KPIs in near real time using a typical 15-minute sampling interval, and autonomously execute optimized policies \cite{jiang2025survey}. Through reflection and learning loops, they continuously refine their operational strategies, adapting to changing network conditions and usage dynamics. Therefore, as presented in \cite{jiang2025tutorial}, improving the intelligence level of automation in RAN management and optimization is one of the core application scenarios of Agentic AI. By leveraging LAMs as autonomous agents, networks gain capabilities in decision-making, collaborative task execution, and self-learning. These agents enhance automation, fault diagnosis, and resource management across edge and 6G networks, including Digital Twin environments. The expected impact will be higher in next-generation Open Radio Access Network where agents could analyze real-time load and demands to autonomously fine-tune resource allocation, boosting performance, resilience, and reliability, especially given Open RAN’s modular architecture and support for open interfaces.

\subsection{Industry Applications and Impact} 

Besides these future expectations, leading telecom operators and vendors have just started to invest in Agentic AI for RAN, with several initiatives demonstrating its applicability across operational objectives. Deutsche Telekom and Google Cloud have developed a network agent based on Gemini 2.0 to detect anomalies and initiate corrective actions \cite{dt2024}. Telenor and Ericsson have introduced a system that adjusts RAN configurations to balance throughput and energy use, supporting sustainable 6G readiness \cite{ericsson2025}. NVIDIA’s BubbleRAN incorporates Large Telecom Models (LTMs) to support reasoning over topology, KPIs, and policies, contributing to automated monitoring, reconfiguration, and fault handling \cite{nvidia2024}. These developments show how Agentic AI can operate over domain-specific data and protocols to support context-aware decision-making.


\subsection{Challenges of Agentic AI in Mobile Networks}

Deploying Agentic AI in 5G and 6G mobile networks introduces technical and operational challenges. A primary concern is the computational and energy cost of LAMs, particularly in 6G scenarios where inference is distributed across edge and cloud nodes. Solutions under investigation include lightweight LLMs, collaborative inference, and modular reasoning architectures \cite{chen2024mobileai}.

Security and privacy risks are amplified in Agentic systems that process sensitive user data and control signals. Adversarial behaviors, hallucinated outputs, and data leakage require safeguards such as sandboxing, zero-trust protocols, and anomaly detection \cite{acharya2025}.

Efficient communication and coordination protocols are also needed. Natural language interactions among agents, common in current frameworks, are insufficient for high-frequency tasks. Structured, semantically aligned communication is required, consistent with the shift toward semantic-native communication in 6G \cite{jiang2025survey}. The Model Context Protocol (MCP) has been proposed as a semantic orchestration interface linking agents to network functions across RAN, Core, and Service layers, supporting distributed inference and coordinated behavior \cite{kumar2024}.

Agentic systems must also maintain long-term contextual awareness, requiring memory infrastructure, adaptive retrieval, and planning modules. Techniques such as Retrieval-Augmented Generation (RAG) and memory routing support coherent decision-making over time \cite{habib2025}.

Processing multimodal input in real time demands effective resource allocation and QoS control aligned with system goals \cite{chen2024mobileai}. At scale, Agentic AI must operate across heterogeneous environments with open standards, federated orchestration, and compatibility with telecom protocols \cite{sapkota2025}.

Addressing these challenges will require advances in AI architecture, coordination protocols, and system governance.

\section{Agentic AI for 5G RAN optimization: case study}\label{sec:casestudy}

We present a case study that illustrates the application of Agentic AI to the problem of monitoring and interpreting Key Performance Indicators (KPIs) in 5G RAN.
The objective is to launch continuous monitoring processes that can autonomously detect deviations from expected behavior, identify their causes, and propose corrective actions when needed.

\begin{figure}[htbp]
    \centering
    \includegraphics[width=\columnwidth]{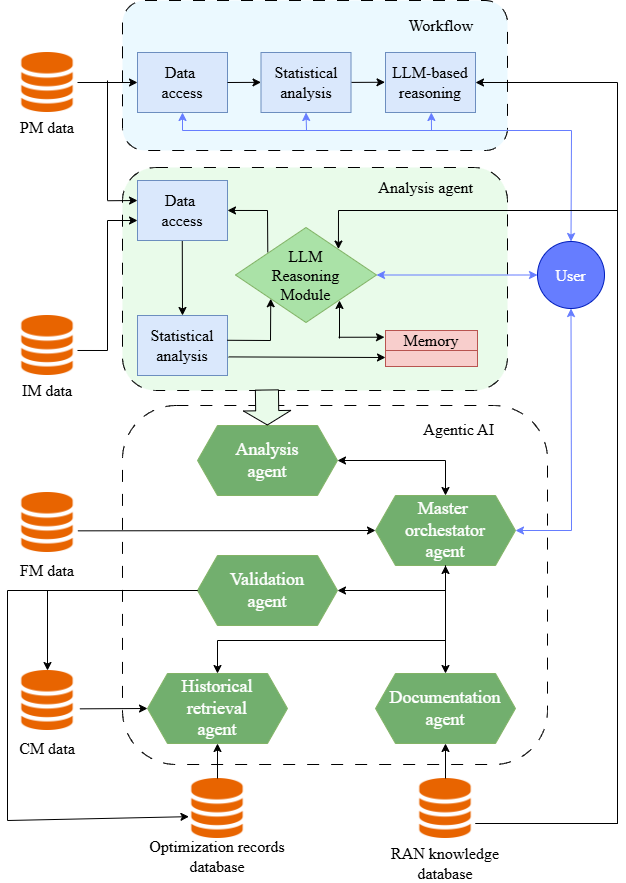}
    \caption{Architecture of an Agentic AI system for RAN management.}
    \label{fig:esquemaglobal}
\end{figure}

As stated in Section~\ref{sec:agenticRAN}, the foundation of Agentic AI systems lies in the richness and variety of data available from the telecom operator. In our example, see Figure~\ref{fig:esquemaglobal}, the primary data source is the ingestion of performance management (PM) data, processed into KPI streams, from the 5G infrastructure. These KPIs streams represent the dynamic state of the 5G RAN. The system is designed to detect deviations from expected patterns, whether anomalies or sustained shifts, and assess their significance.

KPIs can be collected at multiple levels of spatial granularity within the 5G RAN. At the finest granularity, individual cell‐level KPIs capture localized performance. Band- and sector-level KPIs aggregate metrics over sets of cells operating on the same frequency or serving the same area respectively. 
Node-level KPIs summarize performance across all cells controlled by a gNodeB, while regional-level KPIs represent aggregated behavior of network elements within a defined geographical space. At higher level, cluster-level KPIs represent the behavior of grouped network elements.

By spanning this hierarchy—cell, band, sector, node, region, and cluster—the Agentic AI system can reason over both micro- and macro-scale patterns. Combinations across different aggregation levels are also possible, supporting a flexible and context-aware analysis of network performance.

This hierarchical structure enables the system to isolate anomalies at any level and compare peer elements to identify outliers or asymmetrical behaviors.
If the reasoning module determines that the deviation is detrimental and actionable, it can propose and, if permitted, execute an optimization procedure adapted to the affected layer or network elements.

In addition to KPI time-series, as depicted in Figure~\ref{fig:esquemaglobal}, our case study also considers data from other operational sources, including Fault Management (FM), Configuration Management (CM), and Inventory Management (IM) systems. It also incorporates records of past optimization actions, parameter settings, and their observed outcomes. Furthermore, domain knowledge from RAN documentation—such as vendor configuration guides and protocol specifications—is included. These heterogeneous inputs provide the situational context needed for agentic reasoning and informed decision-making.

The next step in the design of our Agentic AI architecture is to incorporate the functional “design patterns” outlined in Section~\ref{sec:patterns}. In particular, the Use Tool pattern structures how each sub-agent interacts with the heterogeneous data sources described earlier. Under this pattern, we explicitly define and encapsulate the following tools:

\begin{itemize}
    \item Data Access Tool retrieves information from underlying data sources by issuing structured queries over the 5G cloud infrastructure.

    \item Statistical Time-Series Analysis Tool applies statistical methods to the data under study, identifying variations across KPI time-series and organizing the results into structured tables that break down the observed changes at multiple RAN hierarchies.

    \item LLM Reasoning Tool is the core reasoning module of our Agentic AI architecture. Large Reasoning Models \cite{li2025} ingest diverse structured inputs—including statistical summaries and other relevant data—and perform reasoning across multiple RAN hierarchies. By analyzing these inputs, the model identifies dependencies, formulates hypotheses.
\end{itemize}

Within the context of a complete agent, the LLM Reasoning Tool must be extended into an LLM Reasoning Module. This expanded component not only executes the analytical tasks defined by the tool but also integrates the functional logic required to support design patterns such as Planning and Reflection. Under the Planning pattern (Section~\ref{sec:planning}),  it decomposes high-level objectives into structured sequences of executable sub-tasks. And under the Reflection pattern (Section~\ref{sec:reflec}), the Agentic AI continually compares its own predictions or intermediate outputs against new data to refine future reasoning steps. In our RAN use case:

\begin{enumerate}
    \item Initial Reasoning Pass: The LLM Reasoning Tool ingests the first set of KPI analysis and hypothesizes why performance deviations occur, how they propagate, and what structural patterns emerge across different levels of the RAN hierarchy.
    \item Self-Reflection Step: The system assesses whether observed anomalies or recurrent behaviors exhibit underlying dependencies across hierarchical layers or among peer elements, thereby establishing relationships between potential causes and their effects within the vertical and lateral structure of the network topology.
    \item Refinement: The reasoning model updates its internal representation of the network state and plans deeper, targeted queries to further investigate the potential root cause. This closed-loop process relies on reasoning interactions that integrate current measurements with previously analyzed data—both from the statistical analysis outputs and from earlier iterations within the reflective cycle itself.
\end{enumerate}

By embedding this Reflective Loop within each reasoning cycle, the Agentic AI reduces blind spots and prevents error propagation. This implementation follows the Reflection pattern described in Section~\ref{sec:patterns}, ensuring that each recommendation is updated based on the most current data.

Finally, under the Multi‐Agent Collaboration Pattern (Section~\ref{sec:collab}) discrete, narrowly scoped agents can work together through clearly defined interfaces to achieve system‐wide goals thus allowing the complete Agentic AI vision.

From the architectural and design principles defined before we designed in Figure~\ref{fig:esquemaglobal} three AI systems with increasing level of intelligence and automation: Workflow, AI Agent and Agentic AI. We only developed the workflow, while for the AI Agent and Agentic AI we only give some illustrative possible operations as described below.

\subsection{Workflow RAN Analysis}

The workflow system corresponds to the most basic instantiation of the proposed architecture, where reasoning is tightly structured and predominantly relies on predefined modular components. In this configuration, the RAN analysis process is sequentially organized around a small set of functional tools operating under the Use Tool pattern.

The process initiates with the Data Access Tool, which retrieves KPI time-series data across RAN hierarchies. This data is then passed to the Statistical Time-Series Analysis Tool, which processes the measurements to detect variations, trends, and anomalies. The outcomes are structured into tables, summarizing deviations at different RAN layers and facilitating structured analysis.

Next, the LLM Reasoning Tool ingests the outputs from the statistical layer along with additional RAN knowledge data. The model performs its first pass of reasoning to hypothesize possible root causes and propagation patterns of observed degradations. A reflective step follows, wherein the model cross-compares its intermediate conclusions with previously seen data and contextual patterns, evaluating consistency across hierarchies and peer elements.

In the final step, the LLM updates its internal network state and formulates additional queries aimed at refining its hypotheses. This iterative process integrates new data with prior outputs, closing the loop between statistical analysis and reasoning. While the Planning and Reflection patterns are not fully implemented in this workflow system, the reasoning steps already initiate partial decomposition of tasks and basic feedback logic.

The modular design and linear composition in this workflow serve as a foundation for more advanced systems. It enables traceable and interpretable behavior, making it suitable for controlled evaluations and early-stage deployment scenarios.

\subsection{AI Agent}\label{sec:agent}

The AI agent builds upon the workflow structure by embedding its components within a cohesive entity capable of autonomous operation. It integrates the same core tools—data access and statistical analysis—but replaces the stateless LLM Reasoning Tool with the stateful LLM Reasoning Module. This module serves as the decision-making core of the agent, maintaining contextual awareness and enabling closed-loop reasoning.

The memory enables retention of intermediate outputs and past reasoning steps, which are used in subsequent inferences. The agent continuously refines its understanding through recurrent invocations of the reasoning module, enabling it to recognize dependencies, revisit prior conclusions, and generate focused follow-up queries using additional IM data. The planning and reflection patterns are natively supported, allowing the agent to decompose high-level tasks into sub-tasks and to adjust its internal representation based on updated data.

Through this structure, the AI agent transitions from a fixed sequence of operations to a semi-autonomous system that adapts to its inputs.

\subsection{Agentic AI}

Agentic AI represents the highest level of automation and intelligence relying on the orchestration of specialized agents interact through well-defined interfaces, implementing planning, reflection, and multi-agent collaboration patterns. Each agent is assigned a specific operational role and operates over distinct data domains to support complex decision-making processes grounded in current and historical network knowledge.

\textbf{Master Orchestrator Agent}: Supervises the multi-agent process, issues coordination directives, and manages inter-agent dependencies. It accesses FM data to track relevant events and generates the final system response.

\textbf{Analysis Agent}: Inherits the capabilities of the AI agent described in Section~\ref{sec:agent}. It performs KPI-based diagnostics by integrating PM and IM data. Its LLM Reasoning Module formulates hypotheses, detects performance degradations, and triggers investigation sequences.

\textbf{Historical Retrieval Agent}: Responsible for retrieving precedent cases, optimization outcomes, or past resolution patterns. It operates over the Optimization Records Database and fetches data from the current network configuration, providing contextual grounding and comparative baselines for ongoing decision tasks.

\textbf{Documentation Agent}:  Interprets RAN knowledge from a theoretical perspective and analyzes vendor and operator documents to establish operational semantics and terminology.

\textbf{Validation Agent}: Upon user approval, applies the modified configuration parameters and initiates an evaluation phase. It retains the previous configuration to enable rollback if degradation is detected; otherwise, it confirms the changes and records the new state.

This modular composition ensures that the system remains flexible, traceable, and adaptable across dynamic network conditions. Together, their collaboration enables efficient and goal-driven RAN management, completing the operational loop from intent expression to validated policy enactment.

\section{Conclusion}\label{sec:conclu}

Agentic AI offers a prospective direction for addressing the growing complexity of mobile communication systems, particularly within the context of 5G and upcoming 6G networks. This paper positions Agentic AI as a structured approach for integrating cognitive mechanisms
into network management systems. The focus lies on autonomously managing and optimizing dynamic environments like the Radio Access Networks, where conventional methods no longer suffice. By drawing parallels between Agentic AI and earlier autonomous agent frameworks, the work establishes a foundation for understanding how these systems can assume active roles in goal-driven network operations.

The applicability of Agentic AI is discussed in the context of RAN management, focusing on its ability to handle changing network conditions and make decisions. The discussion also addresses the challenges involved in scaling these solutions for broader deployment in future infrastructures.

This is examined through a practical case study involving a 5G RAN scenario, using time-series data and language model-based agents to support autonomous control. This example illustrates how different Agentic AI designs can facilitate RAN monitoring and adjustment processes.

\section*{Acknowledgments}
This work has been developed in collaboration with Telefónica, whose support and insights have contributed to the progress of our research group and project activities.


\bibliographystyle{IEEEtran}
\bibliography{refs.bib}


\section*{Biography Section}
\textbf{Jorge Fellejero Fernández} currently pursuing his Ph.D. in Telecommunications engineering at Universidad Politécnica de Madrid, Spain.  

\textbf{Luis Mendo Tomás} received his Ph.D. in Telecommunications engineering at Universidad Politécnica de Madrid (UPM), Spain, in 2001. He is associate professor at UPM.    

\textbf{Zoraida Frías Barroso} received her Ph.D. in Telecommunications engineering at Universidad Politécnica de Madrid (UPM), Spain, in 2015. She is associate professor at UPM.    

\textbf{Luis A. Hernández Gómez} received his Ph.D. in Telecommunications engineering at Universidad Politécnica de Madrid (UPM), Spain, in 1982. He is associate professor at UPM.


\end{document}